\documentclass[11pt,journal,compsoc]{IEEEtran}
\ifCLASSOPTIONcompsoc
  \usepackage[nocompress]{cite}
\else
  \usepackage{cite}
\fi
\usepackage[linesnumbered,ruled,vlined]{algorithm2e}
\usepackage{graphicx}
\usepackage{subcaption}
\usepackage{amsmath,amssymb,amsfonts}
\usepackage{multirow}
\usepackage[utf8]{inputenc}
\usepackage[T1]{fontenc}
\ifCLASSINFOpdf
\else
\fi
\begin{document}
\title{Fuzziness-based Spatial-Spectral Class Discriminant Information Preserving Active Learning for Hyperspectral Image Classification}
\author{Muhammad Ahmad
\IEEEcompsocitemizethanks{\IEEEcompsocthanksitem Department of Computer Engineering, Khwaja Fareed University of Engineering and Information Technology, Rahim Yar Khan, 64200, Pakistan.\protect\\
E-mail: mahmad00@gmail.com}
\thanks{Manuscript received May 28, 2020.}}
\markboth{ArXiv}%
{Ahmad \MakeLowercase{\textit{et al.}}: Bare Demo of IEEEtran.cls for ArXiv}
\IEEEtitleabstractindextext{%
\begin{abstract}
Traditional Active/Self/Interactive Learning for Hyperspectral Image Classification (HSIC) increases the size of the training set without considering the class scatters and randomness among the existing and new samples. Second, very limited research has been carried out on joint spectral-spatial information and finally, a minor but still worth mentioning is the stopping criteria which not being much considered by the community. Therefore, this work proposes a novel fuzziness-based spatial-spectral within and between for both local and global class discriminant information preserving (FLG) method. We first investigate a spatial prior fuzziness-based misclassified sample information. We then compute the total local and global for both within and between class information and formulate it in a fine-grained manner. Later this information is fed to a discriminative objective function to query the heterogeneous samples which eliminate the randomness among the training samples. Experimental results on benchmark HSI datasets demonstrate the effectiveness of the FLG method on Generative, Extreme Learning Machine and Sparse Multinomial Logistic Regression (SMLR)-LORSAL classifiers.
\end{abstract}
}
\maketitle
\IEEEdisplaynontitleabstractindextext
\IEEEpeerreviewmaketitle
\IEEEraisesectionheading{\section{Introduction}\label{sec:introduction}}

\IEEEPARstart{H}{yperspectral Imaging} (HSI) is concerned with the extraction of meaningful information form the objects of interest-based on the radiance acquired by the sensor from long, medium or short distance \cite{Ahmad2019A}. HSI technology has been investigated in a wide variety of urban, mineral exploration, environmental, in-depth classification of forest areas, monitoring the pollution in city areas, investigating the coastal and domestic water zones, inspection for natural risks, i.e. flood, fires, earthquakes, eruptions \cite{Ding2016, Mahmad2018}.

Therefore, to enhance the applicability, robust, effective and automatic Hyperspectral Image Classification (HSIC) methods are required. The classical kernel-based methods \cite{Melgani2004} are effective and robust. Nevertheless, these methods are inadequate to handle ill-posed conditions \cite{Aydemir2017}. Furthermore, it is well established fact that the HSIC performance depends on the quality and size of training data \cite{Deng2018}. However, in the scenario of limited training samples, classical HSIC methods do not perform well \cite{Ahmad2018, Ahmad122}. Therefore, the main objective of this work is to develop a novel method to automatically extract the meaningful information captured by HSI sensors particularly in the case when the labeled training samples are not adequate or not fully reliable \cite{Ahmad2017}. In a nutshell, the following specific contributions are made in this work.

\begin{enumerate}
    \item FLG utilizes the fuzziness concept instead of uncertainty and couples it with the samples diversity while maximizing separability measure using total local and global for both within and between class discriminant information. 
    \item Total global class information impairs the local topology and cannot satisfactorily characterize the local class discriminant information. This may lead to instability of within-class compact representation. 
    \item A similar problem whereby local class discriminant information only considers local between and within-class information and ignore the global class information. 
    \item To overcome the above two problems, we define a novel objective function that jointly considers the total global and local class discriminative information. The proposed discriminative objective function assesses the stationary behavior of samples in the spectral domain in a fine-grained manner.
\end{enumerate}

The objective function proposed in this work is adapted for deriving a set of spatially heterogeneous samples that jointly optimize the above terms. The objective function is based on the optimization of a multi-objective problem for the estimation of within and between class scatters while preserving the total local and global class discriminant information. Thus, FLG significantly increases the generalization capabilities and robustness of classical machine learning classifiers.

The rest of the paper is structured as follows. Section \ref{RW} examines the related works. Section \ref{Methodology} presents the theoretical aspects of \textit{FLG}. Section \ref{Exp} discusses the details of experimental settings. Section \ref{Exp2} provides the details about HSI datasets, experimental results and comparison with the state-of-the-art methods. Finally, section \ref{Con} summarizes the contributions and discuss future research directions.

\section{Related Work}
\label{RW}

In the last decade, kernel-based methods have been successfully applied for HSIC. However, kernel-based methods do not perform well when the ratio between the number of labeled training samples and spectral bands \cite{Ahmad11,  Ahmad113, Ahmad112} is small \cite{Chi2007}. Several alternative methods have been proposed to address the issues related to the limited availability of labeled training samples. One of them is to iteratively enlarge the original training set in an interactive process, i.e, human-human interaction or human-machine interaction \cite{Li2006}.

Active Learning (AL) is an iterative process of selecting informative samples from a set of unlabeled samples. The selection choice is based on a ranking of scores that are computed from a model's outcome. Carefully selected samples are added to the training set and the classifier is retrained with the new training set. The training with selected samples is robust because it uses samples that are suitable for learning. Thus, the sample selection criteria are a key component of the AL framework \cite{Ahmad2019A, ahmad2020}. The most common sample selection criteria are formalized into three different groups.

\begin{enumerate}
    \item Uncertainty of samples \cite{Lewis1994, Settles2010} and query by committee \cite{Seung1992, Tuia2011}

    \item Influence on the model such as length of gradients \cite{Dong2009} and Fisher information ratio \cite{David1996}.
    
    \item Intrinsic structure and distribution of the unlabeled samples such as Gaussian similarity \cite{Zhu2005}, Kullback-Leibler divergence similarity \cite{Andrew1998}, manifold learning \cite{Zhang2011}, clustering \cite{Xu2007} and density-weighting \cite{Burr2008} methods. 
    
    \item AL combined with a special classifier such as AL combined with Logistic Regression \cite{Li2010}, SVM \cite{Huo2014} and Gaussian process regression \cite{Pasolli2012}.
\end{enumerate}

These sample selection methods have shown that the AL process significantly improves the performance of any classifier while querying the informative samples \cite{Rubens2015}. However, in HSIC, the collection and labeling of queried samples are associated with a high cost in terms of time. Therefore, most of the previous studies focused on selecting the single sample in each iteration (stream-based), by assessing its uncertainty \cite{Mitra2004}. This can be computationally expensive because the classifier has to be retrained for each new labeled sample.

Pool-based (Batch-mode) methods have been proposed to address the above-mentioned issues by assessing the pool of samples. The major drawback of multiclass pool-based models is that the pool of selected samples brings redundancy i.e., no new knowledge or information is provided to the classifier in the retraining phase.

This work addresses the above-mentioned issues by defining a multiclass fuzziness-based total local and global for both within and between-class scatter information preserving AL pipeline. This work explicitly considers spatial-spectral heterogeneity of the selected samples by defining a novel discriminative objective function and properly generalize it. The combination of the above criteria results in the choice of the potentially most informative and heterogeneous samples than ever possible. The proposed method is experimentally compared with state-of-the-art methods and based on the comparisons, some guidelines are derived to use AL techniques for HSIC.

\section{Methodology}
\label{Methodology}

A number of sample selection methods for AL have been proposed in the literature \cite{Ahmad2018, Li2013, Luo2004, Shi2015} though uncertainty \footnote{Most uncertain sample has similar posteriori probability for two possible classes.}-based sample selection remains popular due to its simplicity. The probabilistic classification models can directly be used to compute the uncertainty whereas it's not that simple for non-probabilistic models \cite{Yu2018}.

Let us assume a HSI cube can be represented as $X = [x_1, x_2, x_3, \dots, x_L]^T \in \mathcal{R}^{L\times(M \times N)}$ composed of $(M\times N)$ samples per band belonging to $C$ classes and $L$ bands  \cite{Ahmad2017C, Ahmad2016, AhmadPK17}. Further assume that $(x_i, y_j)$ be a sample of Hyperspectral cube in which $y_j$ is the class label of $x_i$ sample. We first randomly select $n$ number of labeled training samples to form a training set $X_T$ and rest has been selected for the test set $X_V$. We further make sure that $n \ll m$ and $X_T \cap X_V = \emptyset$ \cite{Ahmad2019A} for each iteration of our proposed AL method.

\subsection{Fuzziness}
\label{Fuz}
A probabilistic/non-probabilistic classifier produces the output $\mu = \mu_{ij}$ of $m \times C$ matrix containing probabilistic/non-probabilistic outputs. There is no need to compute the marginal probabilities for probabilistic classifiers however it is mandatory for non-probabilistic classifiers and for that discriminative random field is used to compute the probabilities. These probabilistic outputs are used to construct a membership matrix which must satisfy the properties \cite{Ahmad2019A, Ahmad2018} $\sum_{j = 1}^{C} \mu_{ij} = 1 ~~ and ~~ 0 < \sum_{i = 1}^{N} \mu_{ij} < 1$ where $\mu_{ij} = \mu_j(x_i) \in [0,1]$ represent the membership of $x_i$ to the $y_j$ class. Later this membership matrix is used to compute the fuzziness of $m$ samples for $C$ class as;

\begin{multline}
E(\mu) = \frac{-1}{C} \sum_{i=1}^{N}\sum_{j = 1}^{C} [\mu_{ij}log(\mu_{ij}) + \\ 
(1 - \mu_{ij})log(1-\mu_{ij})]
\label{Eq2}
\end{multline}
\subsection{Fuzziness Categorization}
\label{Ca}

We first make a matrix of fuzziness information associated with samples' spatial information and their predicted and actual class labels and a test set, i.e., $[E(\mu),~ X_V(s),~ y_j(a),~ y_j(p),~ X_V]$ where $X_V(s)$ stands for spatial information of test set, $y_j(a)$ and $y_j(p)$ represents the actual and predicted class labels. Later a median-based fuzziness categorization method is proposed to select the samples to compute the class scatter information. There are two ways to compute the median value to make $\mathcal{F}_1$ and $\mathcal{F}_2$ sets depending on the number of total samples. If the total number of samples is odd then the median can be computed as:

\begin{equation}
    \mathcal{M}(\mathcal{F}_1) = \frac{m_1+1}{2}
\end{equation}

\begin{equation}
    \mathcal{M}(\mathcal{F}_2) = \frac{m_2+1}{2}
\end{equation}

If the total number of samples in the test set is even then the median can be computed as;

\begin{equation}
    \mathcal{M}(\mathcal{F}_1) = \frac{\frac{m_1}{2} + \frac{m_1+1}{2}}{2}
\end{equation}

\begin{equation}
    \mathcal{M}(\mathcal{F}_2) = \frac{\frac{m_2}{2} + \frac{m_2+1}{2}}{2}
\end{equation}
where $m_1$ and $m_2$ refers to the total number of samples in $\mathcal{F}_1$ and $\mathcal{F}_2$ respectively. Now we have two sets of fuzziness then we will find the median values in both sets and keep these values in \(\mathcal{Q}_1\) and \(\mathcal{Q}_2\). By this process, we create two fuzziness groups and place the samples in low ($[0-0.5]$ fuzziness magnitude) and high ($[0.5-1.0]$ fuzziness magnitude) fuzziness groups, respectively. From these two sets we select misclassified foreground samples to compute the local and global class discriminate information to select the heterogeneous spectral samples for the training set.

\subsection{Global Class Discriminant Information}
\label{Global}

The fuzziness categorization process returns $\mathcal{X} = \{(\textbf{\textit{x}}_i~ | ~ i = 1, \dots, \mathcal{K} ~ and ~ y_j ~ | ~ j = 1, \dots, C)\}$, where $\textbf{\textit{x}}_i \in \mathcal{R}^{L \times (M \times N)}$, $\mathcal{K} < m$ and $n \ll \mathcal{K}$. The global class discriminant information preserving process aims to attain a space $\mathcal{R}^{L \times (M \times N)}$ in which each training sample $\textbf{\textit{x}}_i$ can be well represented by $\textbf{\textit{x}}_i\rightarrow y_j \in \mathcal{R}^{L \times (M \times N)}$. Therefore, linear discriminant analysis (LDA) is most efficient method to preserve the global information in computational efficient fashion. LDA seeks a linear projection matrix $\textbf{U} \in \mathcal{R}^{L \times (M \times N)}$ that maximizes the Fisher discriminant ratio (FDR) as follows:

\begin{equation}
	\textbf{U} = \underset{\textbf{U}} {\mathrm{argmax}} \frac{|\textbf{U}^T S_B \ \textbf{U}|}{|\textbf{U}^T S_W \ \textbf{U}|}
	\label{Eq10}
\end{equation}
where $S_B$, $S_W \in \mathcal{R}^{L*L}$ are the global between and within class scatter matrices respectively.

The performance of FDR is highly dependent on the quality of the scatter matrices i.e. when the number of samples in each class is much smaller than the number of bands, $S_W$ would be singular. This makes the inverse of $S_W$ and the eigenvalue decomposition $S_W^{-1} S_B$ impracticable which is commonly known as small sample size problem \cite{Gao2012}. However, non-probabilistic LDA (NPLDA) \cite{Sharma2015} and probabilistic vector-based LDA (PLDA) \cite{Prince2012} methods have been proposed to address the problems. In NPLDA, regularized LDA (RLDA) use PCA to reduce the dimensionality of HSI from $L$ to $L^* < L$ and performs the classical LDA. Therefore, the size of within-class scatters matrix $S_W$ is reduced from $L \times L$ to $L^* \times L^*$, making $S_W$ well-conditioned in the PCA subspace as long as $L^*$ is small enough. Although PCA$+$LDA solves the singularity problem with the expense to lose geospatial information due to the lossy compression of PCA.

RLDA can hardly extract meaningful information when the size of the training samples becomes large. Unlike NPLDA counterparts, PLDA solves the singularity problem by capturing between and within-class variation under the probabilistic framework. Whereas, NPLDA extracts meaningful information by manipulating the scatter matrices. While the PLDA avoids the inverse of ill-conditioned $S_W$ through probabilistic modeling of within and between class variation, individually. Moreover, since it explicitly characterizes both the class and noise components, PLDA has a unique advantage in capturing discriminative information. It may be discarded or considered as less important by its NPLDA counterparts \cite{Prince2012}.

Since NPLDA and PLDA successfully address the singularity problem of LDA, however, these are still far from solving the small sample size problem. In real-life, many HSI datasets are in the form of tensors and tensor structures are useful in alleviating the small sample size problem. Vectorization or reshaping consequently breaks the valuable tensor structures. To cope with these issues, multi-linear LDA (MLDA) approaches have been proposed to gain more robustness. According to different criteria used in projection learning, MLDA can be grouped into two categories, i.e. ratio-based MLDA (RMLDA) \cite{Lu2009} and difference-based MLDA \cite{Tao2008}.

RMLDA consider $\mathcal{X} = \{\{\textbf{\textit{x}}_{ij} \in \mathcal{R}^{L_c * L_r}\}^{n}_{i=1}\}^{C}_{j=1}$ as a set of true class HSI samples obtained by fuzziness process. The aim of RMLDA is to find projections that maximize the ratio between and within class scatter. For instance, $2D$-RMLDA learn two matrices $U_r \in \mathcal{R}^{L_r * q_r}$  and $U_s \in \mathcal{R}^{L_s * q_s}$, which characterize the column and row spaces respectively. These matrices solved alternately based on the following scatter ratio criterion. By fixing $U_r$, $U_s$ is solved by:

\begin{equation}
	U_c = \underset{U} {\mathrm{argmax}} \frac{tr(U^T {{S_B}_s} U)}{tr(U^T {{S_W}_s}  U)}
	\label{Eq11}
\end{equation}
where ${S_W}_s \in \mathcal{R}^{L_s * L_s}$ and ${S_B}_s \in \mathcal{R}^{L_s * L_s}$ are the MLDAs within and between class scatter matrices respectively. These matrices are further decomposed as follows;

\begin{equation}
	{S_B}_s = \sum_{j=1}^{C} N_j (M_j - M) U_r U^T_r (M_j - M)^T
	\label{Eq12}
\end{equation}

\begin{equation}
	{S_W}_s = \sum_{ij} (x_{ij} - M_j) U_r U^T_r (x_{ij} - M_j)^T
	\label{Eq13}
\end{equation}
where $N_j$ be the total number of samples in $j^{th}$ class, $M = \frac{1}{N} \sum_{ij} \textbf{\textit{x}}_{ij}$ is the overall mean matrix and $M_j = \frac{1}{N_j} \sum_{i=1}^{N_j} \textbf{\textit{x}}_{ij}$ is the class mean matrix. Analogous to LDA, the solution $U_s$ is given by the eigenvectors of ${S_W}_s^{-1}{S_B}_s$ associated with the $q_s$ largest eigenvalues. By fixing $U_s$, the solution of the row projection $U_r$ can be obtain similarly. Similar to LDA, the above objective functions (Equations \ref{Eq12} and \ref{Eq13}) are further be decomposed into the following two objective functions:

\begin{equation}
	S_{GB} = \sum_{j=1}^C N_j U^T(M_j - M) (M_j - M)^T U
	\label{Eq14}
\end{equation}

\begin{equation}
	S_{GW} = \sum_{j=1}^C \bigg( \sum_{i=1}^{K+1} U^T (x_{ij} - M_j) (x_{ij} - M_j)^T U \bigg)
	\label{Eq15}
\end{equation}

Equation \ref{Eq15} aims to ensure that the samples from the center are as close as possible if and only if they belong to the same class whereas equation \ref{Eq14} makes sure that the center of each class from the total center is as distant as possible. Furthermore, MLDA aims to find projections that maximize the difference between within and between-class scatter. For instance, MLDA learns multi-linear projections alternately and maximizes the following objective function for the column projection:

\begin{equation}
	U_s = \underset{U} {\mathrm{argmax}} \bigg(U^T ({S_B}_s - \xi {S_W}_s) U \bigg)
	\label{Eq16}
\end{equation}
where $\xi$ is the tuning parameter which is heuristically set to the largest eigenvalue of $({S_W}_s)^{-1}{S_B}_s$. After a simple derivation, the solution $U_s$ is given by the eigenvectors of ${S_B}_c - \xi {S_W}_c$ associated with the $q_s$ largest eigenvalues. By exploiting the tensor structures, MLDA can learn more reliable multi-linear scatter matrices that have smaller sizes and much better conditioning than LDA. MLDA solves the singularity problem of LDA and gains robustness in estimating global class discriminant information with a small sample size \cite{Luo2009}. Equation \ref{Eq16} have a shortcoming, for instance, it may impair the local topology but cannot satisfactorily characterize the local class discriminant information of HSI data. This may lead to instability of within-class compact representation. To overcome these difficulties, this work explicitly computes the local class discriminant information.

\subsection{Local Class Discriminant Information}
\label{Local}

We incorporate the geometrical transformation method which preserves the local class discriminant information by building two adjacency matrices i.e. intrinsic and penalty matrix. The intrinsic matrix characterizes class compactness and connects each sample with its neighboring samples of the same class. While the penalty matrix connects marginal points characterize within class separability. Thus, the objective functions presented in \cite{Ding2015} are redefined as follows.

\begin{equation}
	U^* = \underset{U} {\mathrm{argmax}}  \frac{S_{LB}}{S_{LW}}
	\label{Eq17}
\end{equation}

In which, local within and between class compactness is characterized by the following criteria:

\begin{equation}
	S_{LW} = \sum_{i_{i \in S_{k_1}^W}(j)} \ \sum_{j_{j \in S_{k_1}^W}(i)}  \lVert U^T \textbf{\textit{x}}_i - U^T \textbf{\textit{x}}_j \rVert ^2
	\label{Eq18}
\end{equation}

\begin{equation}
	S_{LB} = \sum_{i_{i \in S_{k_2}^B}(j)} \ \sum_{j_{j \in S_{k_2}^B}(i)}  \lVert U^T \textbf{\textit{x}}_i - U^T \textbf{\textit{x}}_j \rVert ^2
	\label{Eq19}
\end{equation}

\begin{equation}
	S_{LW} = 2 U^T \mathcal{X} (D^{LW} - W^{LW}) \mathcal{X}^T U
	\label{Eq20}
\end{equation}

\begin{equation}
	S_{LB} = 2 U^T \mathcal{X} (D^{LB} - W^{LB}) \mathcal{X}^T U
	\label{Eq21}
\end{equation}
where $S_{k_1}^W (i)$ and $S_{k_2}^W (i)$ indicates the index set of $k_1$ and $k_2$ nearest neighbors of $\textbf{\textit{x}}_i$ in the same class. $D^{LW}$ and $D^{LB}$ are diagonal matrix whose columns or rows are the sum of $W^{LW}$ and $W^{LB}$ because both are symmetric matrices, e.g. $D_{ii}^{LW} = \sum_j W_{ji}^{LW}$ and $D_{ii}^{LB} = \sum_j W_{ji}^{LB}$. Thus, similar to equations \ref{Eq14} and \ref{Eq15}, the equation \ref{Eq17} is decomposed into the following two objective functions.

\begin{equation}
	S_{LW} = U^T \mathcal{X} (D^{LW} - W^{LW}) \mathcal{X}^T U
	\label{Eq22}
\end{equation}

\begin{equation}
	S_{LB} = U^T \mathcal{X} (D^{LB} - W^{LB}) \mathcal{X}^T U
	\label{Eq23}
\end{equation}

Equations \ref{Eq22} and \ref{Eq23} are local within and between-class scatters and can characterize the local topology of the samples. Equation \ref{Eq22} aims to make samples compact if and only if they belong to the same class. Equation \ref{Eq23} aims to make them separable if and only if they belong to the different classes. Equations \ref{Eq22} and \ref{Eq23} have a similar problem whereby they only consider local between and within-class scatters but ignore global class discriminant information. It is worth noting that Equation \ref{Eq17} finds the optimal space by maximizing the ratio among within and between-class scatters and to address it, we rewrite equation \ref{Eq17} as follows.

\begin{equation}
	\hat{U} = \underset{U} {\mathrm{argmax}} \frac{U^T \mathcal{X} (D^{LB} - W^{LB}) \mathcal{X}^T U} {U^T \mathcal{X} (D^{LW} - W^{LW}) \mathcal{X}^T U} 
	\label{Eq24}
\end{equation}

The optimal projection vector $U$ that maximizes equation \ref{Eq24} is given by the solution of the maximum eigenvalue to the generalized eigenvalue as shown in the following.

\begin{equation}
	\mathcal{X} (D^{LB} - W^{LB}) \mathcal{X}^T U = \lambda \mathcal{X} (D^{LW} - W^{LW}) \mathcal{X}^T U
	\label{Eq25}
\end{equation}

The effectiveness is still limited such as $S_{LW}$ is singular in many cases and the optimal projection cannot be directly calculated. To overcome this problem, we define a novel objective function that jointly considers the total local and global class discriminative information in the following section.

\subsection{Objective Function}
\label{Objective}

The parameters $\Omega, \lambda, \Psi = 0.5$ is used to make the trade-off among total within class and total between class scatters and control their proportion as follows.

\begin{equation}
	\hat{U} = \underset{U} {\mathrm{argmax}} \ S(U)
	\label{Eq26}
\end{equation}

\begin{equation}
	\hat{U} = \underset{U} {\mathrm{argmax}} \big(\Omega \ S_B(U) - (1 - \Omega) \ S_W(U)\big)
	\label{Eq27}
\end{equation}

\begin{equation}
	S_W(U) = \lambda S_{GW}  + (1 - \lambda) S_{LW}
	\label{Eq28}
\end{equation}

\begin{equation}
	S_B(U) = \Psi S_{GB} + (1 - \Psi) S_{LB}
	\label{Eq29}
\end{equation}

In above equations $\lambda, \Psi, \Omega = 0.5$ and if $\lambda, \Psi = 0$ the equations \ref{Eq28} and \ref{Eq29} reduces to keep only the local within and between class information. Whereas, if $\lambda, \Psi  = 1$ then the equations \ref{Eq28} and \ref{Eq29} will reduce to keep only global within and between-class scatter information. If $\Omega = 0$, equation \ref{Eq27} reduces to the only total between class scatter whereas, $\Omega = 1$ reduces to the only total within-class scatter. Equation \ref{Eq28} impairs the topological structure of samples and the local within class structure which ignores the global information. To remedy these problems we introduce trade-off coefficient $\Omega = 0.5$ to balance both local and global within class information.

\begin{figure}[!ht]
	\centering
	\includegraphics[scale=0.12]{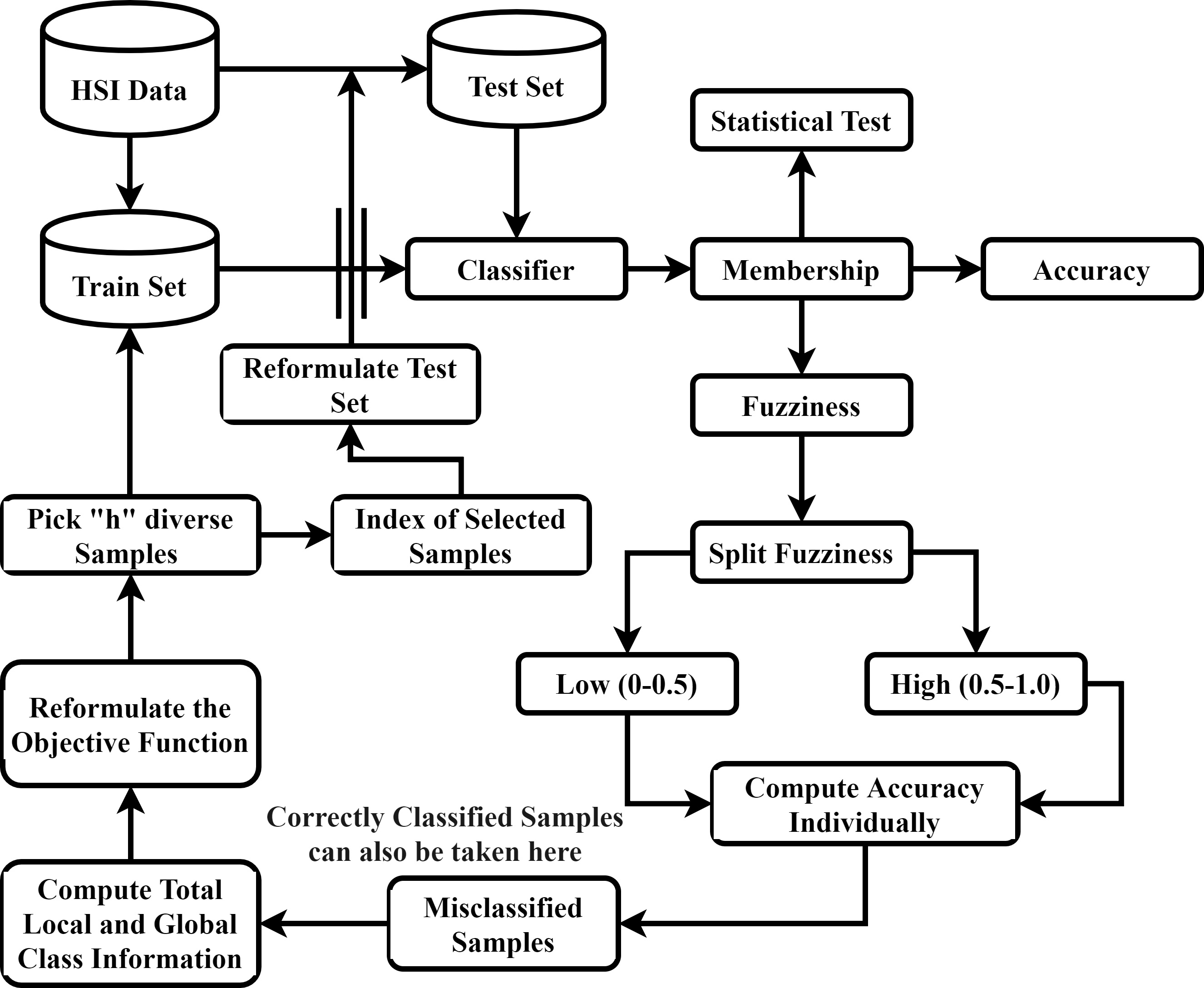}
	\caption{\textit{FLG} Flow-Graph: We first compute the fuzziness and then used a divide and conquer strategy to split the fuzziness values into two groups e.g. Low and High Fuzziness. We pick misclassified samples $(\mathcal{K})$ to compute the global and local class discriminative information and pass this information to the objective function to select $(h \ll \mathcal{K})$ heterogeneous samples.}
	\label{Fig.1}
\end{figure}

The advantages of this integration are to preserve the global and local within class information in which we can achieve total local class discriminant information. Equation \ref{Eq29} involves the global and local between class information, i.e. relationship among sample centers, but ignores local information e.g. relationship among samples if and only if they belong to a different class. However, a single characterization, either global or local between and within-class may be insufficient, which deteriorates classification performance. This work overcomes the above-said limitation by introducing a discriminative objective function, which preserves both global and local between and within-class scatter information as shown in equation \ref{Eq27}. To gain more insight, we solve the equation \ref{Eq27} as follows.

\begin{equation}
	\hat{U} = \underset{U} {\mathrm{argmax}} ({U}^T \ \mathcal{G} \ U)
	\label{Eq36}
\end{equation}
where $\mathcal{G} = \Omega \Psi S_B(U) - (1 - \Omega) \lambda S_W(U) + \mathcal{X}[\Omega (1 - \Psi)S_{LB} - (1 - \Omega)(1 - \lambda)S_{Lw}]\mathcal{X}^T$, in which $D^{LB}$ and $D^{LW}$ are the Laplacian matrices. Thus, the optimal solution of equation \ref{Eq36} is given by $\underset{U} {\mathrm{Max}} \ Tr \{U^T \ \mathcal{G} \ U\}$ with the constraint $U^T U = I$ means the columns of $U$ are orthogonal projection matrix which is able to enhance the discrimination ability among samples and we also used it to reduce the redundancy among the selected samples. Furthermore, we assume that $\mathcal{G}$ is a real symmetric matrix i.e. $\mathcal{G} = \mathcal{G}^T$ because $[\mathcal{X}L^{LB}\mathcal{X}^T]^T = \mathcal{X}L^{LB}\mathcal{X}^T$ and $[\mathcal{X}^T L^{LB}\mathcal{X}]^T = \mathcal{X}^T L^{LB}\mathcal{X}$, moreover,$S_B = S_B^T$ and $S_W = S_W^T$. The complete pipeline of our proposed  method is presented in the Algorithm \ref{ALgo1} and in Figure \ref{Fig.1}.

\begin{algorithm}[!ht]
\footnotesize
\caption{Important steps of our proposed AL method.}
\SetAlgoLined
\KwData{$X_T, X_V$}
\textbf{Initialization:} $\mathcal{X},~\lambda = 0.5,~\Psi= 0.5,~\Omega= 0.5$\;
\While{$|X_T| \leq Threshold$}{
	$\mu_{ij} \leftarrow$ Compute the membership matrix\;
	$E(\mu) \leftarrow$ Compute the fuzziness\;
	$D_V \leftarrow$ Associate the fuzziness, actual and predicted class and spatial information with $X_V$\;
	Categorize the $D_V$ based on Fuzziness into two groups and sort each group in descending order individually\;
	Pick $\mathcal{K}$ misclassified classified samples from each group individually\;
	Compute local and global class information for within class scatter $S_W(U) = \lambda S_{GW}  + (1 - \lambda) S_{LW}$\;
	Compute local and global class information for between class scatter $S_B(U) = \Psi S_{GB} - (1 - \Psi) S_{LB}$\;
	Formulate the objective function $\mathcal{G} = \Omega \Psi S_B(U) - (1 - \Omega) \lambda S_W(U) + \mathcal{X}[\Omega (1 - \Psi)S_{LB} (1 - \Omega)(1 - \lambda)S_{Lw}]\mathcal{X}^T$\;
    Pick $h \ll \mathcal{K}$ most diverse and spectrally heterogeneous samples from $D_V$, add them to $X_T$ and remove from $X_V$\;
\textbf{Repeat until} $|X_T| > Threshold$\;
}
\label{ALgo1}
\end{algorithm}

\section{Experimental Settings}
\label{Exp}

The Hyperspectral datasets used for experimental purposes belong to the wide variety of classification problems in which the number of classes, number of samples and types of sensors are different, i.e, AVIRIS, ROSIS and Hyperion EO-1 Satellite Sensors. The 5-fold-cross-validation process is adopted to evaluate the performance of our proposed pipeline. A necessary normalization between the range of $[0,1]$ is  performed. All the stated experiments are conducted using Matlab $2016b$ installed on a Intel inside Core $i5$ with $8GB$ RAM. 

Experiments have been conducted on several benchmark Hyperspectral datasets using four different types of classifiers, i.e., Support Vector Machine (SVM) \cite{Ahmad2019A}, K Nearest Neighbours (KNN) \cite{Ahmad2019A}, Extreme Learning Machine (ELM) \cite{Ahmad2019A} and MLR-LORSAL \cite{Dopido2012} classifiers. All these classifiers are tuned according to the settings mentioned in their respective works. The first goal of this work is to compare the results of all these classifiers on “High” fuzziness samples while considering the total local and global class discriminant information. Later we compare the results of the best classifier on “Low” fuzziness samples again while considering the total local and global class discriminant information. Finally, the complete pipeline is compared with the state-of-the-art AL methods.

For experimental evaluation, several tests have been conducted including but not limited to Kappa $(\kappa)$, overall accuracy, F1-Score, Precision, and Recall rate. All these evaluation metrics are calculated using the following mathematical formulations. 

\begin{equation}
    OA = \frac{1}{C} \sum_{i = 1}^C TP_i
\end{equation}

\begin{equation}
    \kappa = \frac{P_o - P_e}{1 - P_e}
\end{equation}
where 

\begin{equation}
    P_o = \frac{TP + TN}{TP + FN + FP + TN}
\end{equation}

\begin{equation}
    P_e = P_{Y} + P_{N}
\end{equation}

\begin{multline*}
P_{Y} = \frac{TP + FN}{TP + FN + FP + TN} \times \\ 
        \frac{TP + FN}{TP + FN + FP + TN}
\end{multline*}

\begin{multline*}
P_{N} = \frac{FP + TN}{TP + FN + FP + TN} \times \\ 
        \frac{FN + TN}{TP + FN + FP + TN}
\end{multline*}

where TP and FP are true and false positive, TN and FN are true and false negative computed from the confusion matrix. 

\begin{equation}
    Precision = \frac{1}{C} \sum_{i = 1}^C \frac{TP_i}{TP_i + FP_i}
\end{equation}

\begin{equation}
    Recall = \frac{1}{C} \sum_{i = 1}^C \frac{TP_i}{TP_i + FN_i}
\end{equation}

\begin{equation}
    F1-Score = \frac{2 \times (Recall \times Precision) }{(Recall + Precision)}
\end{equation}

\section{Experimental Results}
\label{Exp2}

In all these experiments, the initial $n = 50$ training samples are selected randomly and the rest of the samples are used as a test example. We start preserving the total local and global class scatter information from these random samples and later used it for the queried samples. In each iteration of AL, $h = 100$ new samples are selected through the proposed pipeline and added back to the original training set.

\subsection{Salinas}
\label{SAL}

\begin{figure}[!hbt]
    \centering
    \begin{subfigure}{0.48\textwidth}
    \centering
    \includegraphics[scale=0.50]{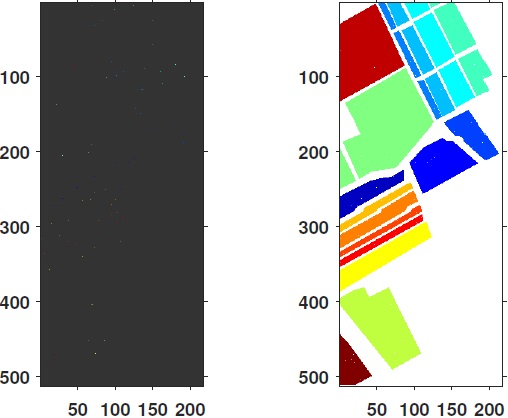}
	\caption{Initial Train and Test Ground Truths without considering the background pixels in training phase.}
    \label{Fig.2A}
    \end{subfigure}
    \begin{subfigure}{0.48\textwidth}
    \centering
    \includegraphics[scale=0.45]{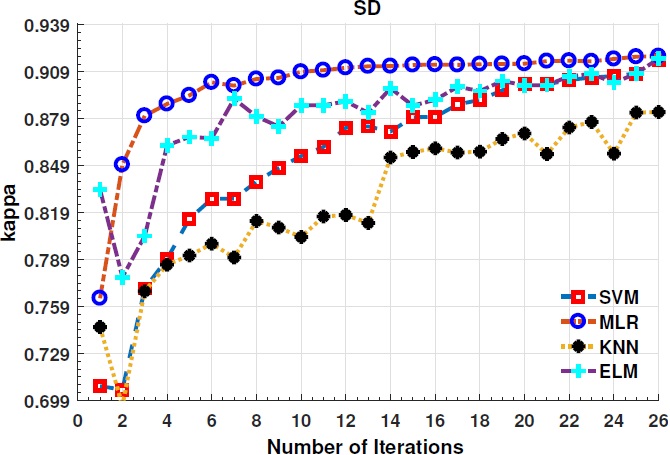}
        \caption{Classification Performance of all classifiers.}
    \label{Fig.2B}
    \end{subfigure}
    \caption{Salinas Dataset (SD): Initial Training and Test Ground Truths (Figure \ref{Fig.2A}) and Classification accuracy with different number of training samples for SVM, KNN, EML and MLR-LORSAL classifiers (Figure \ref{Fig.2B}).}
    \label{Fig.2}
\end{figure}

The Salinas dataset (SD) was acquired over Salinas Valley California using AVIRIS sensor. SD is of size $512\times217\times224$ with a $3.7$ meter spatial resolution with $512\times217$ is spatial and $224$ spectral dimensions. SD consists of vineyard fields, vegetables and bare soils. SD consist of $16$ classes. A few water absorption bands $108-112, 154-167$ and $224$ are removed before analysis. Further details about the dataset can be found at \cite{ahmad2020}. SD class information is provided in Table \ref{Tab.1} and Figure \ref{Fig.2} with the respective class accuracies.

\begin{table}[!hbt]
\centering 
\footnotesize
\caption{SD Class Description i.e., Class name, Test Samples out of which $n = 50$ as initial and $h = [50:100:2500]$ in each iteration, as training samples with their respective overall accuracies for SVM, KNN, ELM and MLR-LORSAL Classifiers over $5$-cross validation.}
    \resizebox{\columnwidth}{!}{
    \begin{tabular}{c|c|c|c|c} \hline
        Class Names  & SVM & KNN & ELM & MLR-LORSAL\\ \hline
        Brocoli green weeds 1 & 0.9654$\pm$0.0514& 0.9576$\pm$0.0555& 0.9917$\pm$0.0101& 0.9831$\pm$0.0127\\ \hline
        Brocoli green weeds 2      & 0.9809$\pm$0.0295& 0.9795$\pm$0.0443& 0.9947$\pm$0.0098& 0.9966$\pm$0.0040\\ \hline
        Fallow                     & 0.9036$\pm$0.1097& 0.8917$\pm$0.1178& 0.9480$\pm$0.0860& 0.9835$\pm$0.0335\\ \hline
        Fallow rough plow          & 0.9166$\pm$0.0617& 0.9883$\pm$0.0117& 0.9871$\pm$0.0126& 0.9956$\pm$0.0055\\ \hline
        Fallow smooth              & 0.9467$\pm$0.0375& 0.9423$\pm$0.0560& 0.9626$\pm$0.0773& 0.9684$\pm$0.0285\\ \hline
        Stubble                    & 0.9587$\pm$0.1074& 0.9918$\pm$0.0058& 0.9855$\pm$0.0503& 0.9863$\pm$0.0295\\ \hline
        Celery                     & 0.9689$\pm$0.0285& 0.9833$\pm$0.0210& 0.9975$\pm$0.0014& 0.9935$\pm$0.0050\\ \hline
        Grapes untrained           & 0.7450$\pm$0.0863& 0.6939$\pm$0.1233& 0.7768$\pm$0.0751& 0.8400$\pm$0.0465\\ \hline
        Soil vinyard develop       & 0.9666$\pm$0.0494& 0.9825$\pm$0.0192& 0.9870$\pm$0.0120& 0.9953$\pm$0.0122\\ \hline
        Corn senesced green weeds & 0.8799$\pm$0.0776& 0.8479$\pm$0.0704& 0.9355$\pm$0.0425& 0.9236$\pm$0.0562\\ \hline
        Lettuce romaine 4wk        & 0.8926$\pm$0.0953& 0.8535$\pm$0.1197& 0.9488$\pm$0.0366& 0.9643$\pm$0.0232\\ \hline
        Lettuce romaine 5wk        & 0.9425$\pm$0.0767& 0.9339$\pm$0.0711& 0.9602$\pm$0.0469& 0.9980$\pm$0.0040\\ \hline
        Lettuce romaine 6wk        & 0.9633$\pm$0.0240& 0.9709$\pm$0.0203& 0.9301$\pm$0.0842& 0.9909$\pm$0.0049\\ \hline
        Lettuce romaine 7wk        & 0.8763$\pm$0.0434& 0.9173$\pm$0.0306& 0.9414$\pm$0.0333& 0.9093$\pm$0.0374\\ \hline
        Vinyard untrained          & 0.6907$\pm$0.0632& 0.5362$\pm$0.1116& 0.6602$\pm$0.1348& 0.6950$\pm$0.0450\\ \hline
        Vinyard vertical trellis   & 0.9157$\pm$0.0875& 0.9181$\pm$0.1101& 0.9718$\pm$0.0218& 0.9421$\pm$0.0783\\ \hline
    \end{tabular}
    }
\label{Tab.1}
\end{table}

\subsection{Indian Pines Dataset}
\label{IPD}
Indian Pines  Dataset (IPD) is obtained over northwestern Indiana’s test site by Airborne Visible / Infrared Imaging Spectrometer (AVIRIS) sensor. IPD is of size $145\times145\times224$ in the wavelength range $0.4-2.5\times10^{-6}$ meters where $145\times145$ is the spatial and $224$ spectral dimensions. IPD consists of $1/3$ forest and $2/3$ agriculture area and other naturally evergreen vegetation. Some corps in the early stages of their growth is also present with approximately less than $5\%$ of total coverage. Low-density housing, building and small roads, Two dual-lane highway and a railway line are also a part of IPD.

\begin{figure}[!hbt]
    \centering
    \begin{subfigure}{0.48\textwidth}
    \centering
    \includegraphics[scale=0.50]{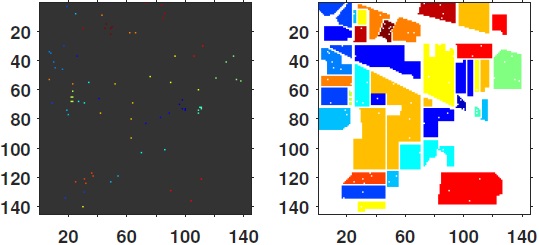}
	\caption{Initial Train and Test Ground Truths without considering the background pixels in training phase.}
    \label{Fig.3A}
    \end{subfigure}
    \begin{subfigure}{0.48\textwidth}
    \centering
    \includegraphics[scale=0.45]{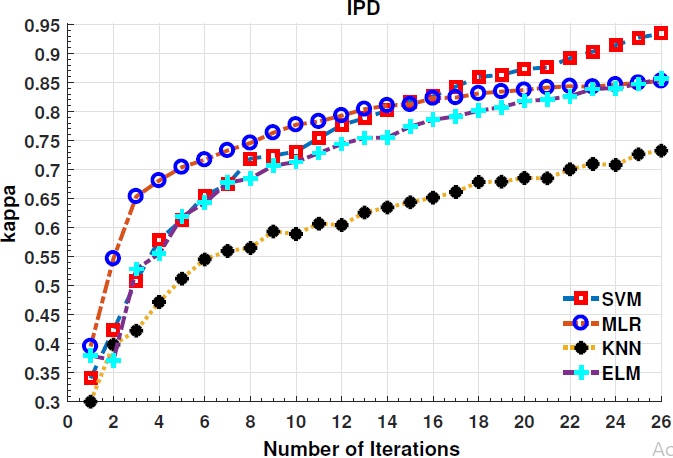}
        \caption{Classification Performance of all classifiers.}
    \label{Fig.3B}
    \end{subfigure}
    \caption{Indian Pines Dataset (IPD): Initial Training and Test Ground Truths (Figure \ref{Fig.3A}) and Classification accuracy with different number of training samples for SVM, KNN, EML and MLR-LORSAL classifiers (Figure \ref{Fig.3B}).}
    \label{Fig.3}
\end{figure}

The IPD ground truth comprised of $16$ classes which are not mutually exclusive. The water absorption bands have been removed before the experiments thus the remaining $200$ bands are used in this experiment. Further details about IPD can be found at \cite{ahmad2020}. IDP class description is provided in Table \ref{Tab.2} and Figure \ref{Fig.3} with the respective class accuracies.

\begin{table}[!hbt]
    \centering
    \footnotesize
    \caption{IPD Class Description i.e., Class name, Test Samples out of which $n = 50$ as initial and $h = [50:100:2500]$ in each iteration, as training samples with their respective $\kappa$ accuracies for SVM, KNN, ELM and MLR-LORSAL Classifiers over $5$-cross validation.}
    \resizebox{\columnwidth}{!}{
    \begin{tabular}{c|c|c|c|c} \hline
        Class Names &  SVM & KNN & ELM & MLR-LORSAL\\ \hline
        Alfalfa & 0.6734$\pm$0.2015& 0.4927$\pm$0.1651& 0.4107$\pm$0.2475& 0.6105$\pm$0.0700\\ \hline
        Corn notill  & 0.7373$\pm$0.1864& 0.5469$\pm$0.0790& 0.7063$\pm$0.1284& 0.7411$\pm$0.1582\\ \hline
        Corn mintill & 0.6899$\pm$0.1809& 0.4650$\pm$0.1370& 0.5666$\pm$0.1569& 0.6674$\pm$0.1549\\ \hline
        Corn & 0.5637$\pm$0.2111& 0.3783$\pm$0.1238& 0.4027$\pm$0.1440& 0.5164$\pm$0.1367\\ \hline
        Grass pasture & 0.8635$\pm$0.1287& 0.7596$\pm$0.1817& 0.8092$\pm$0.1634& 0.8551$\pm$0.0924\\ \hline
        Grass trees & 0.9003$\pm$0.1277& 0.8484$\pm$0.0942& 0.9169$\pm$0.0854& 0.9450$\pm$0.0675\\ \hline
        Grass pasture mowed & 0.8726$\pm$0.1019& 0.7632$\pm$0.2017& 0.4196$\pm$0.1844& 0.8899$\pm$0.0502\\ \hline
        Hay windrowed & 0.9383$\pm$0.0919& 0.8452$\pm$0.1554& 0.9144$\pm$0.1145& 0.9622$\pm$0.0873\\ \hline
        Oats & 0.7069$\pm$0.1460& 0.2724$\pm$0.1430& 0.4190$\pm$0.1302& 0.7502$\pm$0.0722\\ \hline
        Soybean notill & 0.7270$\pm$0.1581& 0.5610$\pm$0.1441& 0.6120$\pm$0.1496& 0.7845$\pm$0.0665\\ \hline
        Soybean mintill & 0.7752$\pm$0.1239& 0.6630$\pm$0.0876& 0.7670$\pm$0.1011& 0.8010$\pm$0.0874\\ \hline
        Soybean clean & 0.6503$\pm$0.2227& 0.3689$\pm$0.1280& 0.6278$\pm$0.2022& 0.6911$\pm$0.1367\\ \hline
        Wheat &  0.9571$\pm$0.0437& 0.8785$\pm$0.1119& 0.9451$\pm$0.0648& 0.9484$\pm$0.0772\\ \hline
        Woods &   0.8795$\pm$0.0967& 0.8668$\pm$0.1122& 0.9141$\pm$0.1045& 0.9419$\pm$0.0565\\ \hline
        Buildings Grass Trees Drives & 0.7082$\pm$0.1496& 0.3949$\pm$0.1245& 0.6854$\pm$0.1820& 0.4994$\pm$0.0842\\ \hline
        Stone Steel Towers & 0.8331$\pm$0.0938& 0.8265$\pm$0.0413& 0.7773$\pm$0.0813& 0.8379$\pm$0.0336\\ \hline
    \end{tabular}
    }
    \label{Tab.2}
\end{table}

\subsection{Kennedy Space Center}

\begin{figure}[!hbt]
    \centering
    \begin{subfigure}{0.48\textwidth}
    \centering
    \includegraphics[scale=0.50]{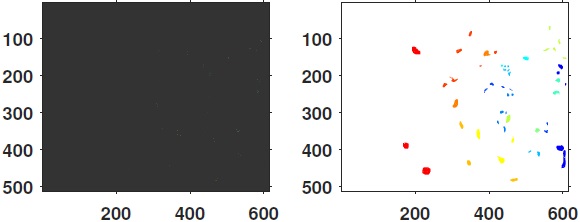}
	\caption{Initial Train and Test Ground Truths without considering the background pixels in training phase.}
    \label{Fig.4A}
    \end{subfigure}
    \begin{subfigure}{0.48\textwidth}
    \centering
    \includegraphics[scale=0.45]{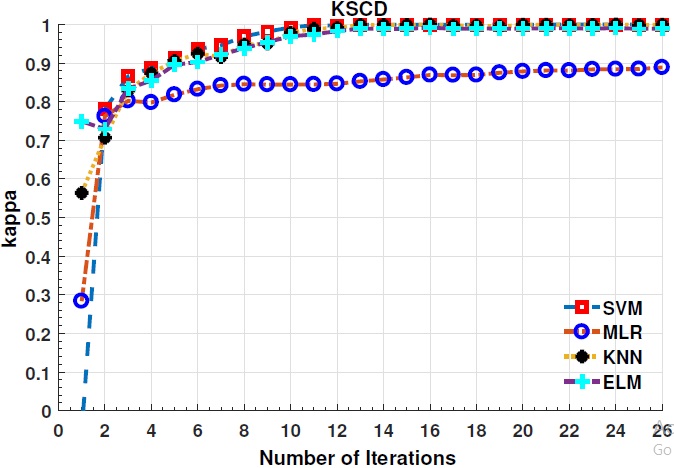}
        \caption{Classification Performance of all classifiers.}
    \label{Fig.4B}
    \end{subfigure}
    \caption{Kennedy Space Center Dataset (KSCD): Initial Training and Test Ground Truths (Figure \ref{Fig.4A}) and Classification accuracy with different number of training samples for SVM, KNN, EML and MLR-LORSAL classifiers (Figure \ref{Fig.4B}).}
    \label{Fig.4}
\end{figure}

Kennedy Space Center dataset (KSCD) acquired on March $23,~ 1996$ using NASA AVIRIS instrument over Kennedy Space Center, Florida. KSCD consists of $224$ spectral bands of $10~nm$ in the wavelength range $400-2500~nm$ with $18$ meter spatial resolution from an altitude of $20~km$. $176$ out of $224$ bands are used for the experimental purposes and remaining water absorption and low SNR bands have been removed. KSCD consist of $13$ classes. KSCD class information is provided in Table \ref{Tab.3} and Figure \ref{Fig.4} with the respective class accuracies.

\begin{table}[!hbt]
    \centering 
    \footnotesize
    \caption{KSCD Class Description i.e., Class name, Test Samples out of which $n = 50$ as initial and $h = [50:100:2500]$ in each iteration, as training samples with their respective $\kappa$ accuracies for SVM, KNN, ELM and MLR-LORSAL Classifiers over $5$-cross validation.}
    \resizebox{\columnwidth}{!}{
    \begin{tabular}{c|c|c|c|c} \hline
        Class Names & SVM & KNN & ELM & MLR-LORSAL\\ \hline
        Scrub           & 0.9422$\pm$0.1953& 0.9692$\pm$0.0588& 0.9588$\pm$0.0742& 0.9305$\pm$0.1339\\ \hline
        Willow Swamp    & 0.9386$\pm$0.1968& 0.9276$\pm$0.1392& 0.9601$\pm$0.0721& 0.9263$\pm$0.0528\\ \hline
        CP/Oak          & 0.9365$\pm$0.1958& 0.9495$\pm$0.1186& 0.9170$\pm$0.1198& 0.8408$\pm$0.1918\\ \hline
        CP hammock      & 0.8864$\pm$0.2254& 0.8785$\pm$0.2113& 0.8075$\pm$0.2008& 0.6052$\pm$0.0952\\ \hline
        Slash Pine      & 0.9002$\pm$0.2101& 0.8701$\pm$0.1955& 0.8428$\pm$0.1902& 0.4006$\pm$0.1145\\ \hline
        Oak/Broadleaf   & 0.8734$\pm$0.2383& 0.8362$\pm$0.2378& 0.7992$\pm$0.1999& 0.5197$\pm$0.0910\\ \hline
        Hardwood Swamp  & 0.9115$\pm$0.2036& 0.9167$\pm$0.1507& 0.9010$\pm$0.1360& 0.7055$\pm$0.1347\\ \hline
        Graminoid Marsh & 0.9312$\pm$0.2034& 0.9253$\pm$0.1729& 0.9385$\pm$0.0869& 0.7754$\pm$0.1620\\ \hline
        Spartina Marsh  & 0.9427$\pm$0.1978& 0.9589$\pm$0.0969& 0.9553$\pm$0.0852& 0.8852$\pm$0.0334\\ \hline
        Cattail Marsh   & 0.9443$\pm$0.1992& 0.9355$\pm$0.1601& 0.9800$\pm$0.0288& 0.8900$\pm$0.1807\\ \hline
        Salt Marsh      & 0.9331$\pm$0.1989& 0.9858$\pm$0.0327& 0.9820$\pm$0.0351& 0.9404$\pm$0.0346\\ \hline
        Mud Flats       & 0.9406$\pm$0.1990& 0.9371$\pm$0.1216& 0.9460$\pm$0.0742& 0.8467$\pm$0.0378\\ \hline
        Water           & 0.9587$\pm$0.1956& 0.9980$\pm$0.0053& 0.9961$\pm$0.0067& 0.9502$\pm$0.1939\\ \hline
 \end{tabular}
 }
 \label{Tab.3}
\end{table}

\subsection{Botswana}
\begin{figure}[!hbt]
    \centering
    \begin{subfigure}{0.48\textwidth}
    \centering
    \includegraphics[scale=0.50]{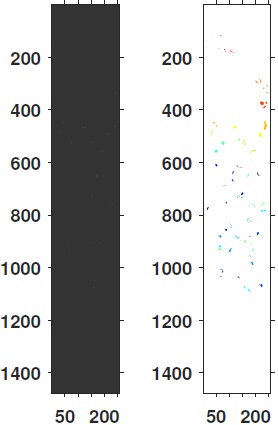}
	\caption{Initial Train and Test Ground Truths without considering the background pixels in training phase.}
    \label{Fig.5A}
    \end{subfigure}
    \begin{subfigure}{0.48\textwidth}
    \centering
    \includegraphics[scale=0.45]{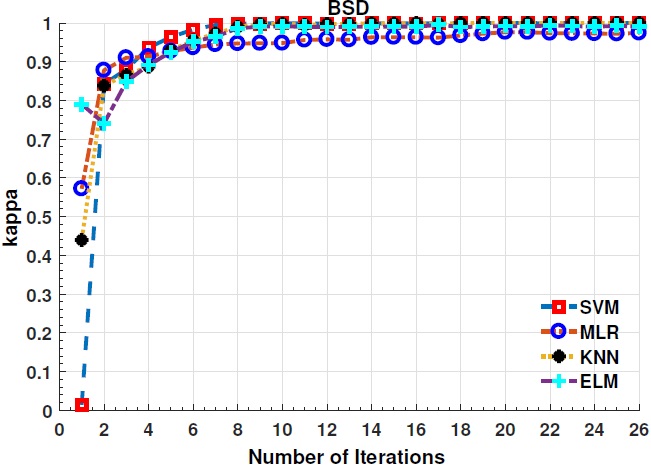}
        \caption{Classification Performance of all classifiers.}
    \label{Fig.5B}
    \end{subfigure}
    \caption{Botswana Dataset (BSD): Initial Training and Test Ground Truths (Figure \ref{Fig.5A}) and Classification accuracy with different number of training samples for SVM, KNN, EML and MLR-LORSAL classifiers (Figure \ref{Fig.5B}).}
    \label{Fig.5}
\end{figure}

Botswana Dataset (BSD) acquired over the Okavango Delta, Botswana on May $31,~ 2001$ through a Hyperion sensor mounted on NASA $EO-1$ Satellite. BSD consists of $242$ bands of $10 nm$ in the wavelength range of $400-2500~nm$ with a $30$ meter spatial resolution from an altitude of $7.7~km$. The number of bands reduced to $145$ from $242$ by removing noisy and uncalibrated bands. The total ground truth classes are $14$ that represent occasional swamps, drier woodlands and seasonal swamps. BSD class information is provided in Table \ref{Tab.4} and Figure \ref{Fig.5} with the respective class accuracies.

\begin{table}[!hbt]
    \centering 
    \footnotesize
    \caption{BSD Class Description i.e., Class name, Test Samples out of which $n = 50$ as initial and $h = [50:100:2500]$ in each iteration, as training samples with their respective $\kappa$ accuracies for SVM, KNN, ELM and MLR-LORSAL Classifiers over $5$-cross validation.}
    \resizebox{\columnwidth}{!}{
    \begin{tabular}{c|c|c|c|c} \hline
        Class Names & SVM & KNN & ELM & MLR-LORSAL\\ \hline
        Water                & 0.9525$\pm$0.1966& 0.9994$\pm$0.0030& 0.9994$\pm$0.0023& 0.9994$\pm$0.0030\\ \hline
        Hippo Grass          & 0.9533$\pm$0.1352& 0.9719$\pm$0.0862& 0.9776$\pm$0.0437& 0.9928$\pm$0.0347\\ \hline
        Floodplain Grasses 1 & 0.9890$\pm$0.0439& 0.9793$\pm$0.0536& 0.9757$\pm$0.0579& 0.9873$\pm$0.0474\\ \hline
        Floodplain Grasses 2 & 0.9544$\pm$0.1954& 0.9660$\pm$0.1174& 0.9742$\pm$0.0672& 0.9773$\pm$0.0741\\ \hline
        Reeds 1              & 0.9419$\pm$0.1982& 0.9362$\pm$0.1421& 0.9517$\pm$0.0824& 0.8883$\pm$0.0965\\ \hline
        Riparian             & 0.9239$\pm$0.2141& 0.8949$\pm$0.2183& 0.9002$\pm$0.1634& 0.8239$\pm$0.1546\\ \hline
        Firescar 2           & 0.9573$\pm$0.1956& 0.9715$\pm$0.1141& 0.9907$\pm$0.0233& 0.9687$\pm$0.1040\\ \hline
        Island Interior      & 0.9581$\pm$0.1956& 0.9672$\pm$0.1026& 0.9758$\pm$0.0555& 0.9837$\pm$0.0771\\ \hline
        Acacia Woodlands     & 0.9494$\pm$0.1955& 0.9371$\pm$0.1510& 0.9464$\pm$0.0902& 0.9298$\pm$0.1032\\ \hline
        Acacia Shrublands    & 0.9474$\pm$0.1968& 0.9577$\pm$0.1026& 0.9264$\pm$0.0876& 0.9464$\pm$0.0648\\ \hline
        Acacia Grasslands    & 0.9419$\pm$0.2000& 0.9624$\pm$0.1109& 0.9664$\pm$0.0609& 0.9283$\pm$0.0960\\ \hline
        Short Mopane         & 0.9502$\pm$0.1955& 0.9684$\pm$0.0934& 0.9632$\pm$0.0723& 0.9369$\pm$0.0137\\ \hline
        Mixed Mopane         & 0.9507$\pm$0.1961& 0.9524$\pm$0.1331& 0.9457$\pm$0.1202& 0.9418$\pm$0.0643\\ \hline
        Exposed Soils        & 0.9229$\pm$0.2124& 0.9948$\pm$0.0141& 0.9789$\pm$0.0396& 0.9776$\pm$0.0847\\ \hline
 \end{tabular}
 }
 \label{Tab.4}
 \end{table}

\subsection{Pavia University}
\begin{figure}[!hbt]
    \centering
    \begin{subfigure}{0.48\textwidth}
    \centering
    \includegraphics[scale=0.50]{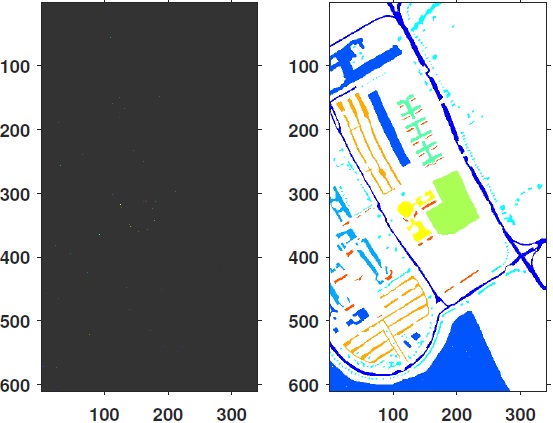}
	\caption{Initial Train and Test Ground Truths without considering the background pixels in training phase.}
    \label{Fig.6A}
    \end{subfigure}
    \begin{subfigure}{0.48\textwidth}
    \centering
    \includegraphics[scale=0.45]{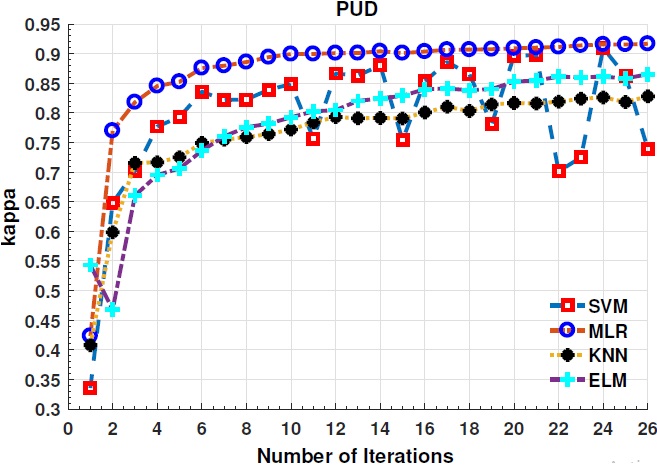}
        \caption{Classification Performance of all classifiers.}
    \label{Fig.6B}
    \end{subfigure}
    \caption{Pavia University Dataset (PUD): Initial Training and Test Ground Truths (Figure \ref{Fig.6A}) and Classification accuracy with different number of training samples for SVM, KNN, EML and MLR-LORSAL classifiers (Figure \ref{Fig.6B}).}
    \label{Fig.6}
\end{figure}

Pavia University Dataset (PUD) gathered over Pavia in northern Italy using a Reflective Optics System Imaging Spectrometer (ROSIS) optical sensor. PUD consists of $610\times610$ spatial and $103$ spectral bands with a spatial resolution of $1.3$ meters. PUD ground truth classes are $9$. PUD class information is provided in Table \ref{Tab.5} and Figure \ref{Fig.6} with the respective class accuracies.

\begin{table}[!hbt]
    \centering 
    \footnotesize
    \caption{PUD Class Description i.e., Class name, Test Samples out of which $n = 50$ as initial and $h = [50:100:2500]$ in each iteration, as training samples with their respective $\kappa$ accuracies for SVM, KNN, ELM and MLR-LORSAL Classifiers over $5$-cross validation.}
    \resizebox{\columnwidth}{!}{
    \begin{tabular}{c|c|c|c|c} \hline
        Class Names & SVM & KNN & ELM & MLR-LORSAL\\ \hline
        Asphalt             & 0.7802$\pm$0.1752& 0.8431$\pm$0.0752& 0.8528$\pm$0.0610& 0.8788$\pm$0.0653\\ \hline
        Meadows             & 0.9082$\pm$0.1252& 0.8942$\pm$0.0646& 0.9072$\pm$0.0765& 0.9537$\pm$0.0904\\ \hline
        Gravel              & 0.6901$\pm$0.0876& 0.6305$\pm$0.1082& 0.6175$\pm$0.1200& 0.7033$\pm$0.0738\\ \hline
        Trees               & 0.8564$\pm$0.1006& 0.8124$\pm$0.0652& 0.8698$\pm$0.0844& 0.8817$\pm$0.0763\\ \hline
        Painted metal sheets & 0.9032$\pm$0.1690& 0.9776$\pm$0.0583& 0.8081$\pm$0.1712& 0.9532$\pm$0.1153\\ \hline
        Bare Soil           & 0.8052$\pm$0.1344& 0.6141$\pm$0.1344& 0.7221$\pm$0.1372& 0.8486$\pm$0.1420\\ \hline
        Bitumen             & 0.7439$\pm$0.1138& 0.6942$\pm$0.0954& 0.6106$\pm$0.1607& 0.8191$\pm$0.0357\\ \hline
        Self-Blocking Bricks & 0.7562$\pm$0.0927& 0.7588$\pm$0.1271& 0.7415$\pm$0.1282& 0.8759$\pm$0.0470\\ \hline
        Shadows             & 0.7950$\pm$0.1790& 0.9981$\pm$0.0018& 0.9945$\pm$0.0101& 0.9972$\pm$0.0060\\ \hline
\end{tabular}
}
 \label{Tab.5}
\end{table}

\begin{table*}[bt]
    \centering
    \caption{Statistical Significance of our proposed Pipeline.}
    \resizebox{\textwidth}{!}{
    \begin{tabular}{c|c|c|c|c|c|c|c|c|c|c|c|c} \hline 
        \multirow{2}{*}{Dataset} & \multicolumn{3}{c|}{SVM} & \multicolumn{3}{c|}{KNN} & \multicolumn{3}{c|}{ELM} &  \multicolumn{3}{c}{MLR-LORSAL} \\
        & Recall & Precision & F1-Score & Recall & Precision & F1-Score & Recall & Precision & F1-Score & Recall & Precision & F1-Score \\ \hline
        SD & 0.9071$\pm$0.0114&0.9213$\pm$0.0133&0.9122$\pm$0.0119 & 0.8993$\pm$0.0166&0.9009$\pm$0.0158&0.8978$\pm$0.0159 & 0.9362$\pm$0.0122&0.9387$\pm$0.0122&0.9359$\pm$0.0119 & 0.9478$\pm$0.0101&0.9443$\pm$0.0089&0.9454$\pm$0.0093 \\ \hline
        KSCD & 0.9261$\pm$0.0046&0.9612$\pm$0.0056&0.9243$\pm$0.0046  & 0.9299$\pm$0.0087&0.9306$\pm$0.0082&0.9282$\pm$0.0079 & 0.9219$\pm$0.0122&0.9271$\pm$0.0113&0.9210$\pm$0.0109 & 0.7859$\pm$0.0288&0.8062$\pm$0.0233&0.7877$\pm$0.0255 \\ \hline
        IPD & 0.7798$\pm$0.0161&0.7715$\pm$0.0174&0.7679$\pm$0.0158 & 0.6207$\pm$0.0284&0.6357$\pm$0.0280&0.6074$\pm$0.0269 & 0.6809$\pm$0.0272&0.7720$\pm$0.0222&0.6986$\pm$0.0245 & 0.7776$\pm$0.0201&0.7546$\pm$0.0192&0.7529$\pm$0.0181\\ \hline
        BSD & 0.9495$\pm$0.0036&0.9862$\pm$0.0018&0.9484$\pm$0.0023 & 0.9614$\pm$0.0042&0.9627$\pm$0.0045&0.9606$\pm$0.0040 & 0.9623$\pm$0.0052&0.9661$\pm$0.0052&0.9629$\pm$0.0047 & 0.9487$\pm$0.0075&0.9471$\pm$0.0067&0.9467$\pm$0.0065 \\ \hline
        PUD & 0.8043$\pm$0.0256&0.8150$\pm$0.0328&0.7980$\pm$0.0273 & 0.8025$\pm$0.0325&0.8077$\pm$0.0328&0.8021$\pm$0.0320 & 0.7916$\pm$0.0317&0.8147$\pm$0.0326&0.7974$\pm$0.0312 & 0.8791$\pm$0.0209&0.8852$\pm$0.0200&0.8799$\pm$0.0191\\ \hline
    \end{tabular}
    }
    \label{Tab.6}
\end{table*}


Here we enlist the experimental results obtained in each iteration of the proposed AL framework for four different types of classifiers, i.e., SVM, KNN, ELM, and MLR-LORSAL. These classifiers have been rigorously used in the literature for comparative analysis. The comparative performance of our proposed AL pipeline using the aforementioned classifiers has been shown in Figures \ref{Fig.2}-\ref{Fig.6}. The tuning parameters of all the above-said classifiers have been explored very carefully in the first few experiments and chosen those which provide the best accuracy. To avoid bias, all the listed experiments are carried out in the same settings on the same machine. Before the experiments, we performed the necessary normalization between $[0, 1]$ and all the experiments are carried out using Matlab $2016b$ installed on an Intel inside Core $i5$ with $8GB$ RAM.

\begin{figure}[!hbt]
    \centering
    \begin{subfigure}{0.23\textwidth}
    \centering
    \includegraphics[scale=0.23]{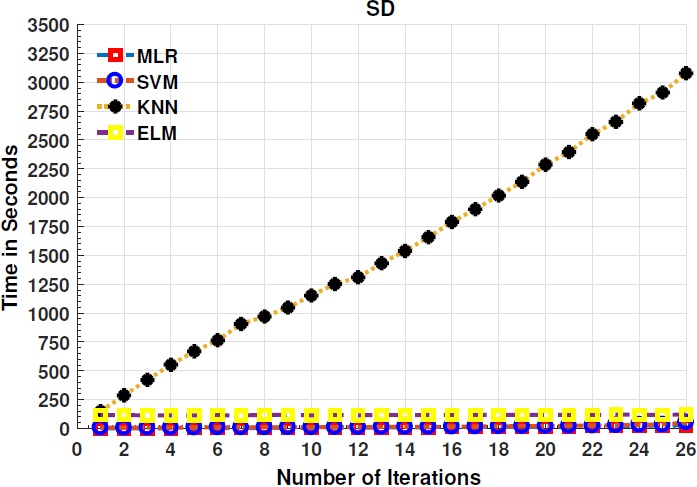}
	\caption{SLD}
    \label{Fig.7A}
    \end{subfigure}
~
    \begin{subfigure}{0.23\textwidth}
    \centering
    \includegraphics[scale=0.23]{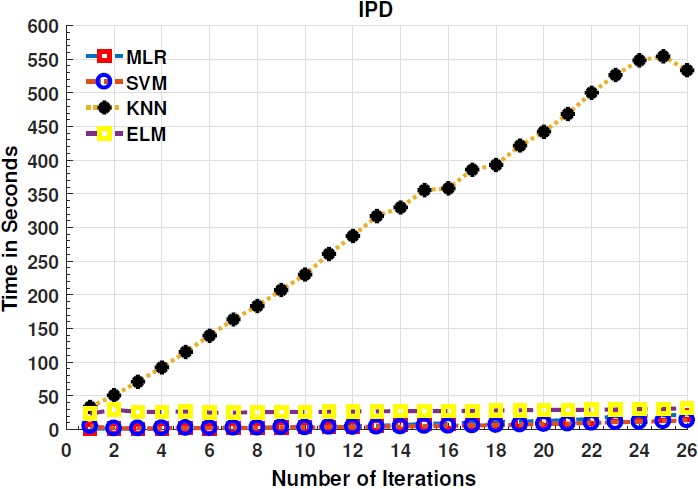}
        \caption{IPD}
    \label{Fig.7B}
    \end{subfigure}
    \begin{subfigure}{0.23\textwidth}
    \centering
    \includegraphics[scale=0.23]{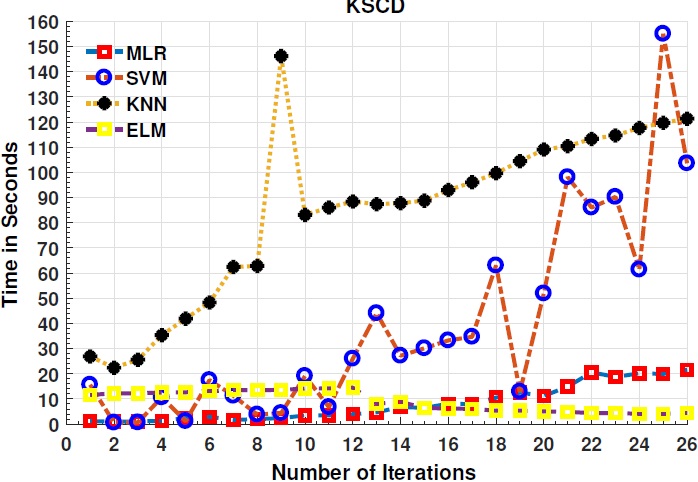}
	\caption{KSCD}
    \label{Fig.7C}
    \end{subfigure}
~
    \begin{subfigure}{0.23\textwidth}
    \centering
    \includegraphics[scale=0.23]{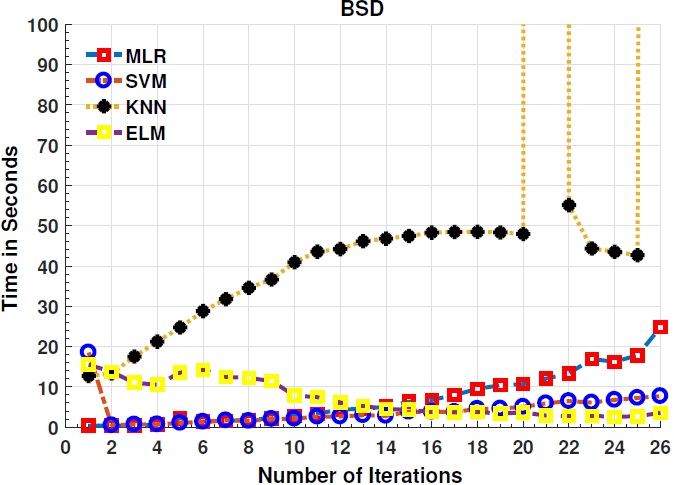}
	\caption{BSD}
    \label{Fig.7D}
    \end{subfigure}
    \begin{subfigure}{0.23\textwidth}
    \centering
    \includegraphics[scale=0.23]{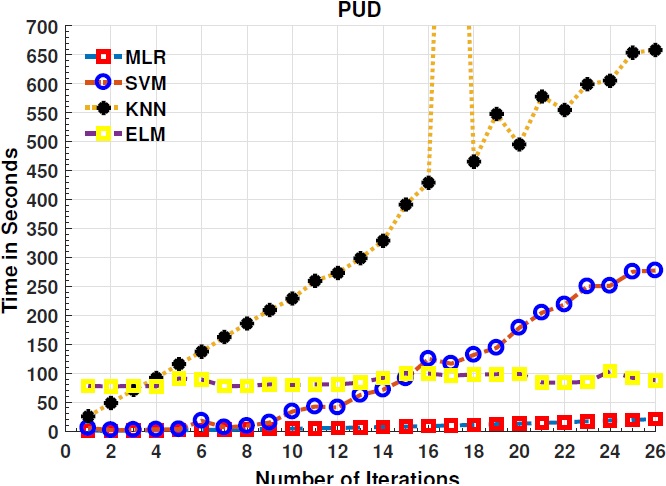}
	\caption{PUD}
    \label{Fig.7E}
    \end{subfigure}
    \caption{Computational Time for all classifiers for each dataset used in this work.}
    \label{Fig.7}
\end{figure}

Here we also enlisted the computational time for SVM, KNN, ELM and MLR-LORSAL classifiers in Figure \ref{Fig.7}. One can observe that the computational time is gradually increasing as the number of training samples increases for all classifiers except for the KNN classifier which exponentially increases. However, the trend is quite different for accuracy which increases exponentially for all classifiers as compared to the time. Computational complexity can significantly reduce for KNN classifiers while using any optimization methods. In our case, we retrain KNN classifier for $k = [2-20]$ in each iteration, which can be overcome using a grid-search type method. All other classifiers have less computational cost and better accuracies, however, SVM and MLR-LORSAL have higher generalization performance then ELM followed by KNN.

The results shown in the above Figures and Tables are based on $5$ Monte Carlo runs with a different number of training samples with equal class representation and in each iteration, the training set size is increased with $h = 100$ of selected samples by our proposed pipeline. It is perceived form Figures and Tables that by including the samples back to the training set, the classification results are significantly improved for all the classifiers and datasets. Moreover, it can be seen that all the classifiers are robust except KNN and their generalization has been significantly increased as shown in Table \ref{Tab.6}.

In this work, we start evaluating our hypotheses from $n = 50$ number of randomly selected labeled training samples and we demonstrate that adding more samples back into the training set significantly increases the accuracy. It is worth noting from experiments that the classifiers trained with selected samples produce better accuracy and improve the generalization performance on those samples which were initially misclassified. To experimentally observe a sufficient quantity of labeled training samples for each classifier, we evaluated the hypotheses with a different number of labeled training samples. Based on the experimental results we conclude that $h = 400-600$ samples obtained by FLG are good enough to produce the acceptable accuracy for Hyperspectral Image Classification.

\section{Conclusion}
\label{Con}
In this paper, a fuzziness-based total local and global class discriminant information preserving active learning method is proposed for HSIC. In this line of investigation, a classifier is trained with a very small set of labeled training samples and evaluated on a large number of unlabeled samples. From results, we observe that it is enough to create a classification model from a small sample instead of a complex model with too many labeled training samples and parameters. Moreover, a small portion of unlabeled samples selected from high fuzziness group to train the model can enhance the generalization performance of any classifier.

The classification results obtained with a different number of labeled training samples prove that the selection of initial labeled training samples does not affect FLG labeling success and does not influence the final classification accuracies. The experimental results show that the FLG which exploits both labeled and unlabeled samples information performs better than standard methods that use only uncertainty information, especially with small sample sizes.
\ifCLASSOPTIONcaptionsoff
  \newpage
\fi
\bibliographystyle{IEEEtran}
\bibliography{sample}
\end{document}